\documentclass{article}

\usepackage{arxiv}

\usepackage[utf8]{inputenc} 
\usepackage[T1]{fontenc}    
\usepackage{hyperref}       
\usepackage{url}            
\usepackage{booktabs}       
\usepackage{amsfonts}       
\usepackage{nicefrac}       
\usepackage{microtype}      
\usepackage{lipsum}
\usepackage{graphicx}
\graphicspath{ {./images/} }

\usepackage{cite}
\usepackage{breqn}
\usepackage{multirow}
\usepackage{eucal}
\usepackage{amsmath,amssymb,amsfonts}

\title{Style-Restricted GAN: Multi-Modal Translation with Style Restriction Using Generative Adversarial Networks}

\author{
 Sho Inoue\\
  Department of Information \& Communication Sciences\\
  Sophia University\\
  \texttt{s-inoue-tgz@eagle.sophia.ac.jp} \\
   \And
 Tad Gonsalves \\
  Department of Information \& Communication Sciences\\
  Sophia University\\
  \texttt{t-gonsal@sophia.ac.jp} \\
}

\DeclareUnicodeCharacter{2212}{-}
\begin{document}
\maketitle
\begin{abstract}
Unpaired image-to-image translation using Generative Adversarial Networks (GAN) is successful in converting images among multiple domains. Moreover, recent studies have shown a way to diversify the outputs of the generator. However, since there are no restrictions on how the generator diversifies the results, it is likely to translate some unexpected features. In this paper, we propose Style-Restricted GAN (SRGAN) to demonstrate the importance of controlling the encoded features used in style diversifying process. More specifically, instead of KL divergence loss, we adopt three new losses to restrict the distribution of the encoded features: batch KL divergence loss, correlation loss, and histogram imitation loss. Further, the encoder is pre-trained with classification tasks before being used in translation process. The study reports quantitative as well as qualitative results with Precision, Recall, Density, and Coverage. The proposed three losses lead to the enhancement of the level of diversity compared to the conventional KL loss. In particular, SRGAN is found to be successful in translating with higher diversity and without changing the class-unrelated features in the CelebA face dataset. To conclude, the importance of the encoded features being well-regulated was proven with two experiments. Our implementation is available at \url{https://github.com/shinshoji01/Style-Restricted\_GAN}.

\end{abstract}



\section{Introduction}
Recent research on Generative Adversarial Networks (GANs)~\cite{GAN}, which are among the most popular generative models, has accelerated and diversified in a variety of directions. One of these is unpaired image-to-image translation, which is a process that allows us to transform an input image into one with minor changes such as sketch-to-photo, opposing gender, or scenery translations. For example, CycleGAN~\cite{cyclegan} and some of its applications can successfully solve these translations via cycle consistency loss (\cite{cyclegan,bicyclegan,singlegan}).

Separately, SingleGAN~\cite{singlegan}, in addition to employing a single generator, enables generator output diversification by introducing an encoder. However, since the diversification experiments are limited to a simple task (coloring changes) using the edge2shoes dataset~\cite{edge2shoes}, its capability to perform complex tasks, such as facial or voice changes, remains unproven. As one example, our investigations encountered some adjustment-related problems when the encoder’s outputs were over-restricted due to the Kullback–Leibler (KL) divergence loss adopted in variational auto-encoders (VAEs)~\cite{VAE}. Furthermore, since diversifications have not been restricted, unexpected characteristics such as background and hair coloring changes occur frequently.

In this paper, we report on a study in which we applied SingleGAN to the translation of facial features with three new losses to compensate for the conventional KL loss. We also show how we pretrained the encoder with the classification task in order to restrict the generator diversifications to class-related features.

\section{Related Work}
\label{related_work}

\subsection{Generative Adversarial Networks}
\label{GAN}

Generative Adversarial Networks (GANs~\cite{GAN}), which are deep generative models, are often compared with VAEs~\cite{VAE}. However, while a VAE tries to reconstruct data exactly as it was input, a GAN learns to generate data that is similar but not necessarily identical to the original input data distribution. A GAN is basically composed of two models, which are referred to as the generator (G) and the discriminator (D). The G generates data in attempts to deceive the D, and the D attempts to determine whether the input data is real or fake.
GAN optimizes the function below:

\begin{equation}
    \begin{split}
    \overset{\text{min}}{_{_G}}\overset{\text{max}}{_{_D}}V(D, G)=\mathbb{E}_{\textbf{x} \sim p_{data}(\textbf{x})}[\log D(\textbf{x})] + \\
    \mathbb{E}_{\textbf{z} \sim \textbf{z})}[\log (1-D(G(\textbf{z}))]
    \end{split}
\end{equation}
\noindent
where the first term indicates the real data judgment and the second shows the generated data. Here, x and z represent the input data and the noise vector, respectively. Although a wide variety of applications have been developed to handle assorted tasks such as cartoon character creation~\cite{cartoon}, photo-realistic image synthesis~\cite{stackgan}, high-resolution human face generation~\cite{pggan}, text-to-image synthesis~\cite{text2image}, and voice conversion~\cite{melgan}, this study focuses on image translation-related tasks.

\subsection{GAN for Unpaired Image-to-Image Translation}
\label{GAN_translation}
Recently, a number of GAN-based image translation studies have been published~\cite{transgaga,pix2pix,nvidia,scribbler,layouts}. Of these, we note that pix2pix~\cite{pix2pix} is capable of supervised image translation using the Conditional GAN concept~\cite{cgan} and paired images. However, since paired images are often unavailable, the utility of this approach is somewhat limited. Unpaired image-to-image translation has also been researched thoroughly, and versions of both DiscoGAN~\cite{discogan} and CycleGAN~\cite{cyclegan} with installed cycle consistency constraints have been proposed. CycleGAN~\cite{cyclegan} adds cycle consistency loss to modified adversarial loss, which enables the model to include original features in images by transforming them back to their original forms. In other words, it makes it possible to restrict transformations within images so that the desired features of the original source images are retained.
This “one-to-one” translation can be extended to “many-to-many” translations using StarGAN~\cite{stargan}, which uses only a single generator and a discriminator by imposing classification tasks on the discriminator. However, this model is not capable of translating to different outputs, which leads to lack of diversification. This problem was laid to rest by some models such as~\cite{munit,stargan2,singlegan}. These models employ an additional module called an encoder for diversified translation. MUNIT~\cite{munit} is the model that separate the style feature into two different perspective: a content code and style code. StarGAN~v2~\cite{stargan2} is the model that diversifies the result by employing a mapping network in addition to the encoder. After the development of StarGAN~\cite{stargan}, SingleGAN~\cite{singlegan}, which we adopted as the base model for our model, was introduced to facilitate additional output diversification.

\subsection{SingleGAN}

Although SingleGAN~\cite{singlegan} allows multi-domain image-to-image translations, including one-to-many, many-to-many, and one-to-one with multi-modal mapping, the translation most relevant to our work is one-to-one with multi-modal mapping, particularly the translation used to diversify the generator output.

In one-to-many translations, in which a generator and discriminators as many as the number of classes are employed, the generator inputs are the source image and the target class, and each discriminator is tasked with determining whether the input image in each class is real or fake. Therefore, when we assume three class translations, the adversarial loss $\mathcal{L}_{adv}$ will be:

\small
\begin{dmath}
    \mathcal{L}_{adv}(D_{\{A,B,C\}}, G) = 
    \sum_{ i,j \in \{A,B,C\},i\neq j} \left( \mathbb{E}_{\chi_i}[\log D_i(x_i)] + 
    \mathbb{E}_{\chi_j}[\log (1-D_i(G(x_j, z_i)))] \right)
\end{dmath}
\normalsize
\noindent
Similarly, the cycle consistency loss $\mathcal{L}_{cycle}$ will be:
\small
\begin{dmath}
    \mathcal{L}_{cycle}(G) = 
    \sum_{i,j \in \{A,B,C\},i\neq j} \left( \mathbb{E}_{\chi_i} \| x_i - G(G(x_i, z_j),z_i) \|_1 \right)
\end{dmath}
\normalsize
\noindent
Where $x_i$ and $z_i$ indicate the input source image and the domain code belonging to class $i$, respectively. Next, for one-to-one translation with multi-modal mapping, the VAE-like encoder is additionally employed to extract the features (latent codes) $c$ related to the “style” of the input images. This style metric indicates, for example, the intensity of emotion or even simply the color of the product (shoes in the original paper~\cite{singlegan}), which is restricted to follow a standard Gaussian distribution $\mathcal{N}(0,I)$ just like a VAE~\cite{VAE}.
Due to the addition of an encoder, two new losses were invented. The first is regression loss ($\mathcal{L}_{reg}$), which is assigned to encourage the generator to capitalize on the latent code. 
\small
\begin{dmath}
    \mathcal{L}_{reg}(G,E) = 
    \mathbb{E}_{\chi_A,\mathcal{N}} \| c_B - E(G(x_A,z_B,c_B), z_B) \|_1
\end{dmath}
\normalsize
\noindent
The second is cVAE loss $\mathcal{L}_{cVAE}$, which is responsible for restricting the latent code with the KL divergence regularization and reconstruction using the encoded feature of the input image and which can be divided into two losses ($\mathcal{L}_{KL}$ and $\mathcal{L}_{cycle'}$).
\small
\begin{dmath}
    \mathcal{L}_{KL}(E) = \mathbb{E}_{\chi} \left[ KL\left(E(x, z) \| \mathcal{N}(0,I)\right) \right]
\end{dmath}
\normalsize
\small
\begin{dmath}
    \mathcal{L}_{cycle'}(G,E) = 
    \mathbb{E}_{\chi_A,\mathcal{N}} \| x_A - G\left(G(x_A,z_B,c_B),z_A,E(x_A,z_A)\right) \|_1
\end{dmath}
\normalsize
\noindent
In cases where c indicates the random vector used for the style selection. These two losses allow models to convert into the image with different styles.

\subsection{Evaluation Method}

In this paper, two evaluation methods were employed to assess the model’s fidelity and diversity, which are essential parts of this research: precision and recall~\cite{pr} and density and coverage~\cite{dc}. Fidelity is evaluated with the precision and density method, while diversity is evaluated with the recall and coverage method. Since both evaluation metrics compare features of the target images to those of the source images, the choice of feature extractors has a significant effect on the process.
In this paper, VGG19~\cite{vgg} with batch normalization~\cite{bn} is adopted as a structure, and its parameters (weights) are defined with three different approaches: random-initialization~\cite{random-initialization}, classification with ImageNet~\cite{imagenet}, and classification with CelebA~\cite{celeba}. Since random-initialization does not limit any condition on the model’s future parameters, it can genuinely reflect the fidelity (diversity) of the outputs. On the other hand, feature extractors pretrained with a classification task are more likely to represent the class-related fidelity (diversity).

\section{Style-Restricted GAN}

As shown in an earlier study, the validity of SingleGAN~\cite{singlegan} was proven with the edge2shoes dataset~\cite{edge2shoes}, which was originally created to convert shoes into different colors from their edges. This coloring task is obviously easier than complex tasks such as changing a facial expression in an image or the emotional content of a voice, primarily because of the need to retain features such as skin/hair color in a facial conversion, so it was necessary to provide a way to boost the model’s diversity while concurrently restricting how it diversifies the result.

\begin{figure}[!htb]
\vskip 0.2in
\begin{center}
\centerline{\includegraphics[scale=0.35]{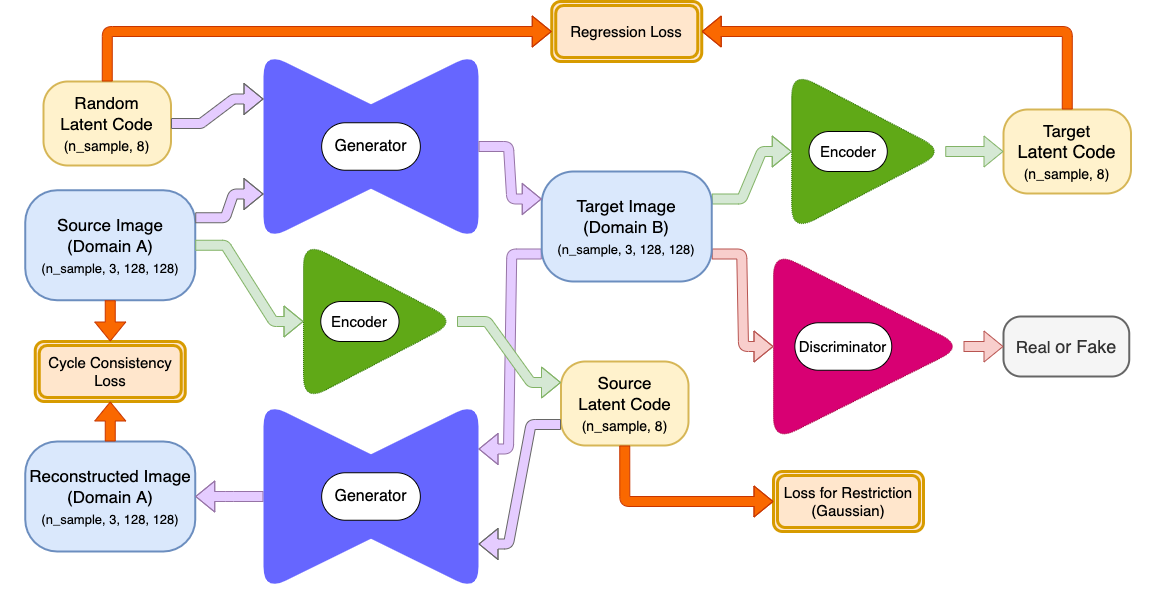}}
\caption{\textit{Overall Flowchart}: It illustrates 3 major losses for this translation and how they are computed. 
Cycle Consistency Loss is the L1 norm between the source image and the reconstructed image using the target image 
and the latent code extracted from the source image.
Regression loss is the L1 norm of input latent code and the latent code encoded using the target image.
Losses for Restriction are computed only from the latent code, which is output from the source image via the encoder.
}
\label{fig:flowchart}
\end{center}
\vskip -0.2in
\end{figure}

Additionally, since the Style-Restricted GAN is designed to perform many-to-many translations with different styles instead of one-to-one translations, 
the class input for the encoder was taken out, and the single discriminator was employed~\cite{stargan}. As for the evaluation metric, cross-entropy loss was replaced by mean square error loss, as in LSGAN~\cite{lsgan}.  
This led to some modifications to the formulation of the $\mathcal{L}_{adv}$, $\mathcal{L}_{cycle}$, and $\mathcal{L}_{reg}$ losses.
Further, it facilitates the implementation of the auxiliary $\mathcal{L}_{class}$ loss. The Style-Restricted GAN flowchart is shown in Fig.~\ref{fig:flowchart}.

\small
\begin{dmath}
    \mathcal{L}_{adv}(D, G) = 
    \sum_{ i,j \in \{A,B,C\},i\neq j} \left( \dfrac{1}{2} \mathbb{E}_{\chi_i}[\left(D_i(x_i)\right)^2] + 
    \dfrac{1}{2} \mathbb{E}_{\chi_j, \mathcal{N}}[\left(D_i(G(x_j,z_i,c_i)-1\right)^2] \right)
\end{dmath}
\normalsize
\small
\begin{dmath}
    \mathcal{L}_{cycle}(G,E) = 
    \sum_{i,j \in \{A,B,C\},i\neq j} \left( \mathbb{E}_{\chi_i, \mathcal{N}} \| x_i - G(G(x_i, z_j, c_j),z_i,E(x_i, z_i)) \|_1 \right)
\end{dmath}
\normalsize
\small
\begin{dmath}
    \label{eq:regression}
    \mathcal{L}_{reg}(G,E) = 
    \sum_{i,j \in \{A,B,C\},i\neq j} \mathbb{E}_{\chi_i,\mathcal{N}} \| c_j - E(G(x_i,z_j,c_j), z_j) \|_1
\end{dmath}
\normalsize
\small
\begin{dmath}
    \mathcal{L}_{class}(D) = 
    \sum_{i,j \in \{A,B,C\}} \dfrac{1}{2} \mathbb{E}_{\chi_i} \left(D(x_i)_{class}-z_i\right) 
\end{dmath}
\normalsize
\noindent

\subsection{Proposed Losses}

One of the major modifications to the proposed model is the introduction of three new losses, “Batch KL Divergence”, “Correlation”, and “Histogram Imitation”, which are designed to more appropriately restrict the distribution of the encoded features. Like the KL divergence loss used in VAE~\cite{VAE}, each loss is responsible for imitating the Gaussian distribution.

\subsubsection{Batch KL Divergence Loss}

Unlike conventional KL divergence loss, which is employed to restrict the distribution of the encoder’s output by referring to the entire dataset, batch KL divergence loss focuses exclusively on the batch limitations. This loss $\mathcal{L}_{bKL}$ can be written as:
\small
\begin{dmath}
    \mathcal{L}_{bKL}(E) = \mathbb{E}_{\chi} \left[ KL\left(E(x_{batch}, z_{batch}) \| \mathcal{N}(0,I)\right) \right]
\end{dmath}
\normalsize
\noindent
where $x_{batch}$ and $z_{batch}$ indicate the input images and the domain codes in the selected batch.

\subsubsection{Correlation Loss}
The following two losses are enhancements of the previously described batch KL divergence loss. The first, correlation loss, is simple. Since there are no correlations among the features in the different dimensions of each data, which is a part of Gaussian distribution, this attribute should remain in the encoder’s output. Therefore, the correlation loss $\mathcal{L}_{corr\_enc}$ will be:
\small
\begin{dmath}
    \mathcal{L}_{corr\_enc}(E) = \mathbb{E}_{\chi} \left[mean|CorrMat(E(x_{batch}, z_{batch})) - I| \right]
\end{dmath}
\normalsize
\noindent
Correlation matrices are computed via a new function called $CorrMat$.

\subsubsection{Histogram Imitation Loss}
\begin{figure}[!htb]
\vskip 0.2in
\begin{center}
\centerline{\includegraphics[scale=0.30]{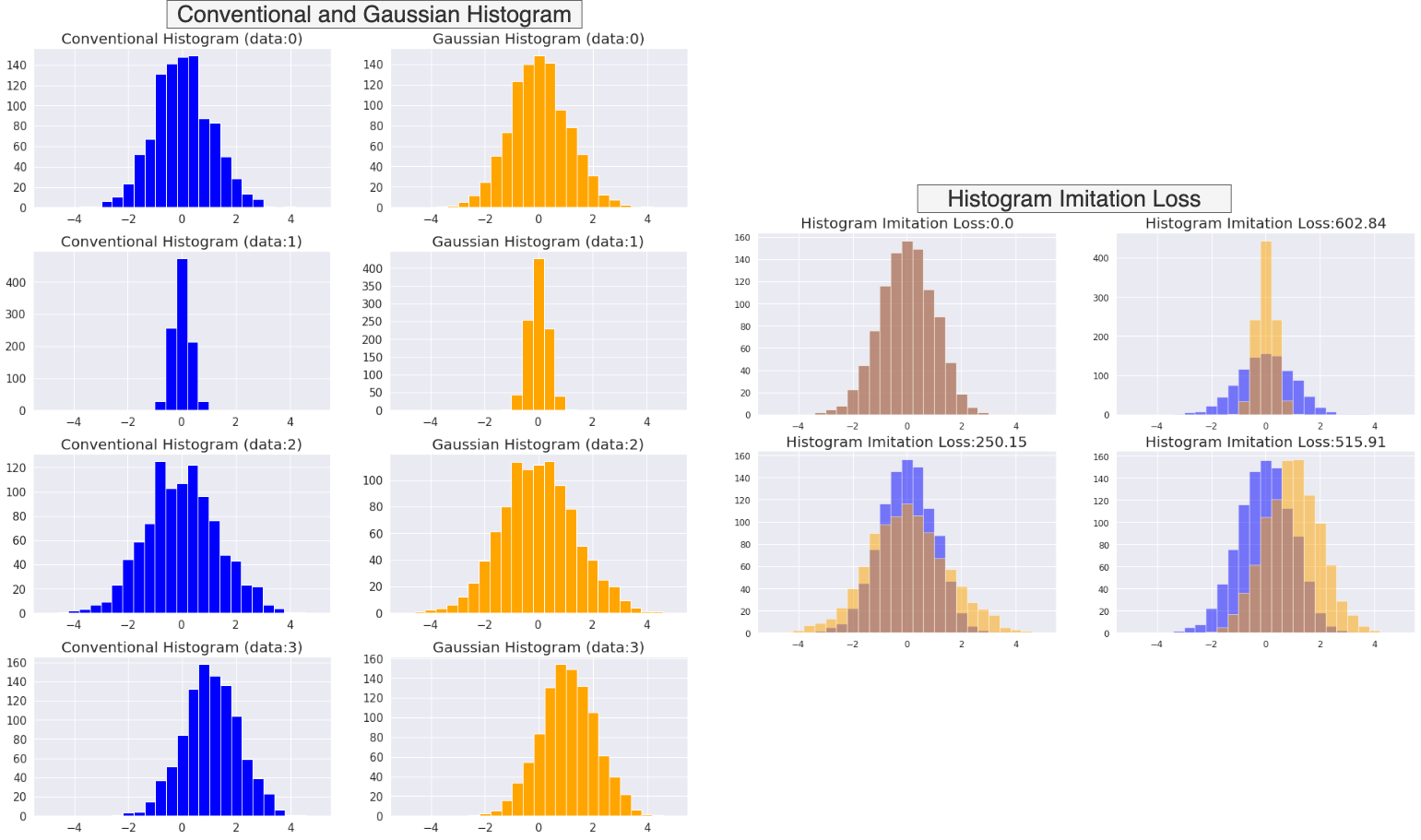}}
\caption{\textit{Gaussian Histogram and Histogram Imitation Loss}:
It shows the comparison between conventional histograms and Gaussian histograms (first 2 columns) and their histogram imitation loss (KL divergence loss) (last 2 columns).}
\label{fig:histogram}
\end{center}
\vskip -0.2in
\end{figure}

In addition to the correlation matrix, the histogram of each feature dimension should be similar to that of Gaussian distribution.
Hence, the histogram imitation loss is designed to make that possible. However, since the computation of a histogram contains a discrete process, it is not reversible, which means that backpropagation process errors will occur. To prevent this, we employ a Gaussian expansion approximation technique called Gaussian Histogram, which is defined as follows:
\small
\begin{dmath}
    \label{eq:delta}
    \Delta w = \dfrac{(max - min)}{bins} 
\end{dmath}
\begin{dmath}
    h_k = \sum_i^{num}\dfrac{1}{\sigma \sqrt{\pi^2}} \exp\left( \dfrac{x_i - \mu_k}{2 \sigma^2} \right) \Delta w
\end{dmath}
\normalsize
\noindent
This process includes numerous variables. First, in Eq.~\ref{eq:delta}, $max$ and $min$ refer to the maximum and minimum value of the histogram and $bins$ indicates the number of divisions within a selected range (histogram classes). Second, $h_k$ and $\mu_k$ show the frequency score of the $k$-th histogram class and its center (histogram class) value. Both histograms (encoded features and Gaussian distribution derived data) are constructed through this computation, and the $\mathcal{L}_{hist}$ histogram imitation loss reflects its similarity using KL divergence loss between the histograms’ distributions.
\small
\begin{dmath}
    \mathcal{L}_{hist}(E) = \mathbb{E}_{\chi} \left[ KL\left(GH(E(x_{batch}, z_{batch})) \| GH(\mathcal{N}(0,I)\right)) \right]
\end{dmath}
\normalsize
\noindent
where $GH$ refers to the Gaussian histogram function. In our experiment, $max$, $min$, $bins$, and $\sigma$ are $10$, $−10$, $50$, and $0.2$, respectively. Fig.\ref{fig:histogram} shows four Gaussian histogram samples, the conventional histogram, and their $\mathcal{L}_{hist}$.

\subsection{Style Restrictions}
\begin{figure}[!htb]
\begin{center}
\centerline{\includegraphics[scale=0.40]{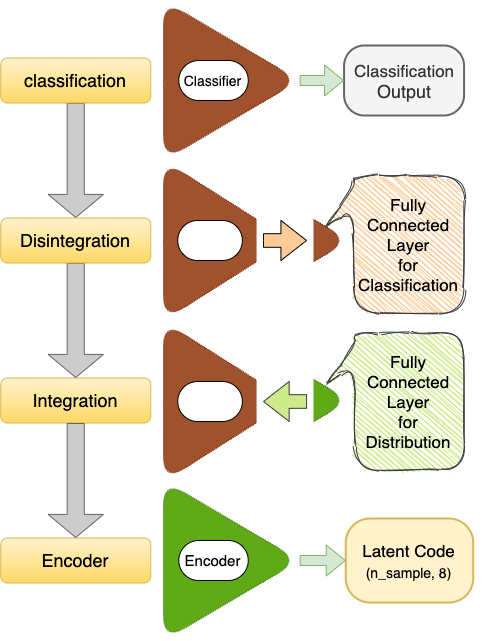}}
\caption{\textit{Flowchart for Style-Restriction}: Style-Restriction is simply composed of 4 stages.  Firstly, a classifier with the same structure as the encoder is learned to classify.  Then, the last layer is taken out and compensated for with a fully connected layer for distribution. Lastly, it is trained to follow Gaussian distribution, with pretrained layers' parameters fixed.}
\label{fig:flowchart_style-restriction}
\end{center}
\vskip -0.2in
\end{figure}

In this paper, we introduce methods to both enhance and restrict model diversity. The model diversification limitations are very important as there are numerous features that we do not want to change, such as the shape of a building in a sketch-to-photo translation, the skin or hair color in a facial expression translation, and the emotional content of a spoken statement in a voice change. We tackle these problems by applying transfer learning to an encoder that is pretrained with classification tasks in the classes used in the translation, as shown in Fig.\ref{fig:flowchart_style-restriction}. The concept behind this process is that a well-trained classifier will eliminate the attributes that do \textit{not} affect the classification. As a result, only the features that are relevant to the classification will remain in the latter process.
Therefore, in our model, we leave out the encoder class input and, for the sake of simplicity, employ the discriminator used in StarGAN~\cite{stargan}, which capitalizes on an auxiliary classifier~\cite{auxiliary} rather than the employment of multiple discriminators like SingleGAN~\cite{singlegan}. Additionally, we adopt the following three techniques to improve our results: UnrolledGAN~\cite{unrolledGAN}, CycleGAN style identity loss ($\mathcal{L}_{idt}$)~\cite{cyclegan}, and regression loss for identity image ($\mathcal{L}_{idt\_reg}$). Finally, the combined losses will be our target loss ($\mathcal{L}_{all}$).

\small
\begin{dmath}
    \mathcal{L}_{idt}(G,E) = 
    \sum_{i \in \{A,B,C\}} \left( \mathbb{E}_{\chi_i} \| x_i - G(x_i, z_i, E(x_i, z_i)) \|_1 \right)
\end{dmath}
\normalsize
\small
\begin{dmath}
    \mathcal{L}_{idt\_reg}(G,E) = 
    \sum_{i \in \{A,B,C\}} \mathbb{E}_{\chi_i,\mathcal{N}} \| c_i - E(G(x_i,z_i,c_i), z_i) \|_1
\end{dmath}
\normalsize
\small
\begin{dmath}
    \label{eq:whole}
    \mathcal{L}_{all} = \mathcal{L}_{adv} + \lambda_{cycle} \mathcal{L}_{cycle} + \lambda_{idt} \mathcal{L}_{idt} + \lambda_{reg}
    \mathcal{L}_{reg} + \lambda_{idt\_reg} \mathcal{L}_{idt\_reg} + \mathcal{L}_{class}
    \lambda_{KL} \mathcal{L}_{KL} + \lambda_{bKL} \mathcal{L}_{bKL}
    + \lambda_{corr\_enc} \mathcal{L}_{corr\_enc} + \lambda_{hist} \mathcal{L}_{hist}
\end{dmath}
\normalsize
\noindent

\subsection{Network Architecture}
The network structure of Style-Restricted GAN (SRGAN) is almost the same as that of SingleGAN~\cite{singlegan}. Similar to some models (\cite{cyclegan,stargan,singlegan}), a generator has two convolutional layers for down-sampling followed by the six-block-ResNet~\cite{resnet} structure with central biasing instance normalization~\cite{cbin} instead of an up-sampling normalization technique before two transposed convolutional layers. Additionally, two PatchGAN discriminators are employed~\cite{patchgan} to determine whether an input is real or fake in different scales with four convolutional layers and no normalization layers. For the multi-model translation encoder, we adopted the same ResNet structure used in BicycleGAN~\cite{bicyclegan}.

\section{Preparation}
\subsection{Dataset}
As mentioned above, the CelebA dataset~\cite{celeba}, which is one of the most popular face datasets, was used for several reasons, including numerous annotation labels and an ample amount of data. The domain class is defined using two characteristics: \textit{male} and \textit{smiling}. Therefore, classes are composed of “male, smiling”, “male, non-smiling”, “female, smiling”, and “female, non-smiling”. To facilitate experiments and focus on the important parts, some images, with or without certain attributes, are excluded, as shown below.
\subsubsection{Characteristics for Classes}
\textit{male}, \textit{smiling}
\subsubsection{Necessary Characteristics}
\textit{No Beard}
\subsubsection{Excluded Characteristics}
\textit{5 o clock shadow}, \textit{blurry}, \textit{chubby}, \textit{double chin}, \textit{eyeglasses}, \textit{goatee}, \textit{mustache}, \textit{sideburns}, \textit{wearing hat}

\subsection{Preprocessing}
In the class conversion process, the image in the dataset is first center-cropped to (178 x 178) and then subsequently resized to (128 x 128). The maximum and the minimum value of each image are converted to 1 and -1, respectively. To augment the dataset, a random horizontal flip is employed in the training process. For the classification task used in the evaluation method, most of preprocessing techniques are the same except for the normalization part, which is normalized to a range of [0, 1] followed by normalization using mean=[0.485,0.456,0.406] and std=[0.229,0.224,0.225], which are separate for each channel.

\subsection{Training Condition}
For classification and translation tasks, the models were trained with Adam optimizer~\cite{adam} set to $\beta_1=0.5$ and $\beta=0.999$. The generator, discriminator, and encoder learning rates were set to $0.0002$, $0.0001$, and $0.001$, respectively, and decreased exponentially as the epoch progressed. This experiment requires numerous hyperparameters, as stated in \ref{eq:whole}, and these parameters are determined referring to the original SingleGAN and our previous experiments, as shown in Table~\ref{table:hyper-params}. Where “or” means it is dependent on the experiments and the models.

\begin{table}[!h]
\caption{\textit{Value of Hyper-Parameters}}
\label{table:hyper-params}
\begin{center}
\begin{small}
\begin{sc}
\renewcommand{\arraystretch}{1.25}
\begin{tabular}{c||c||c}
\toprule
Symbol & Meaning & Value \\
\midrule
$\lambda_{cycle} $ & Cycle Consistency Loss & 5 \\
$\lambda_{idt} $ & Identity Loss & 5 \\
$\lambda_{reg} $ & Regression Loss & 0.5 \\
$\lambda_{idt\_reg} $ & Regression Loss (Identity) & 0 or 0.5 \\
$\lambda_{KL} $ & KL Divergence Loss & 0 or 0.1 \\
$\lambda_{bKL} $ & Batch KL Divergence Loss & 0 or 10 \\
$\lambda_{corr\_enc} $ & Correlation Loss & 0 or 100 \\
$\lambda_{hist} $ & Histogram Imitation Loss & 0 or 100 \\
$k $ & k for UnrolledGAN & 1 or 5 \\
$n\_batch $ & batch size & 128 \\
\bottomrule
\end{tabular}
\renewcommand{\arraystretch}{1}
\end{sc}
\end{small}
\end{center}
\vskip -0.1in
\end{table}

\section{Results}
This paper reports on the results of two experiments: one which compared the conventional KL divergence loss and the combination of the proposed three losses, and the other related to the diversity restriction. Both experiments also included quantitative results using the abovementioned fidelity and diversity methods: precision/recall and density/coverage.

\subsection{Diversified Outputs}
In this experiment, we examined the efficacy of the proposed losses using two SingleGAN conditions. We named the SingleGAN with conventional KL condition and the SingleGAN with proposed losses condition as “Conventional KL Loss” and “Proposed Losses”, respectively. A shortened list of the adjusted hyper-parameters is shown in Table~\ref{table:ex1}.

\begin{table}[h]
\caption{\textit{Hyper-Parameters (Ex.1)}}
\label{table:ex1}
\begin{center}
\begin{small}
\begin{sc}
\renewcommand{\arraystretch}{1.1}
\begin{tabular}{c||cc}
\toprule
\multirow{2}{*}{\centering Symbol} & \multirow{2}{6em}{\centering Conventional KL Loss} & \multirow{2}{5em}{\centering Proposed Losses} \\
& & \\
\midrule
$\lambda_{idt\_reg} $ & 0 & 0 \\
$\lambda_{KL} $ & 0.1 & 0 \\
$\lambda_{bKL} $ & 0 & 10 \\
$\lambda_{corr\_enc} $ & 0 & 100 \\
$\lambda_{hist} $ & 0 & 100 \\
$k $ & 1 & 1\\
\bottomrule
\end{tabular}
\renewcommand{\arraystretch}{1}
\end{sc}
\end{small}
\end{center}
\vskip -0.1in
\end{table}

\begin{figure}[!h]
\vskip 0.2in
\begin{center}
\centerline{\includegraphics[scale=0.3]{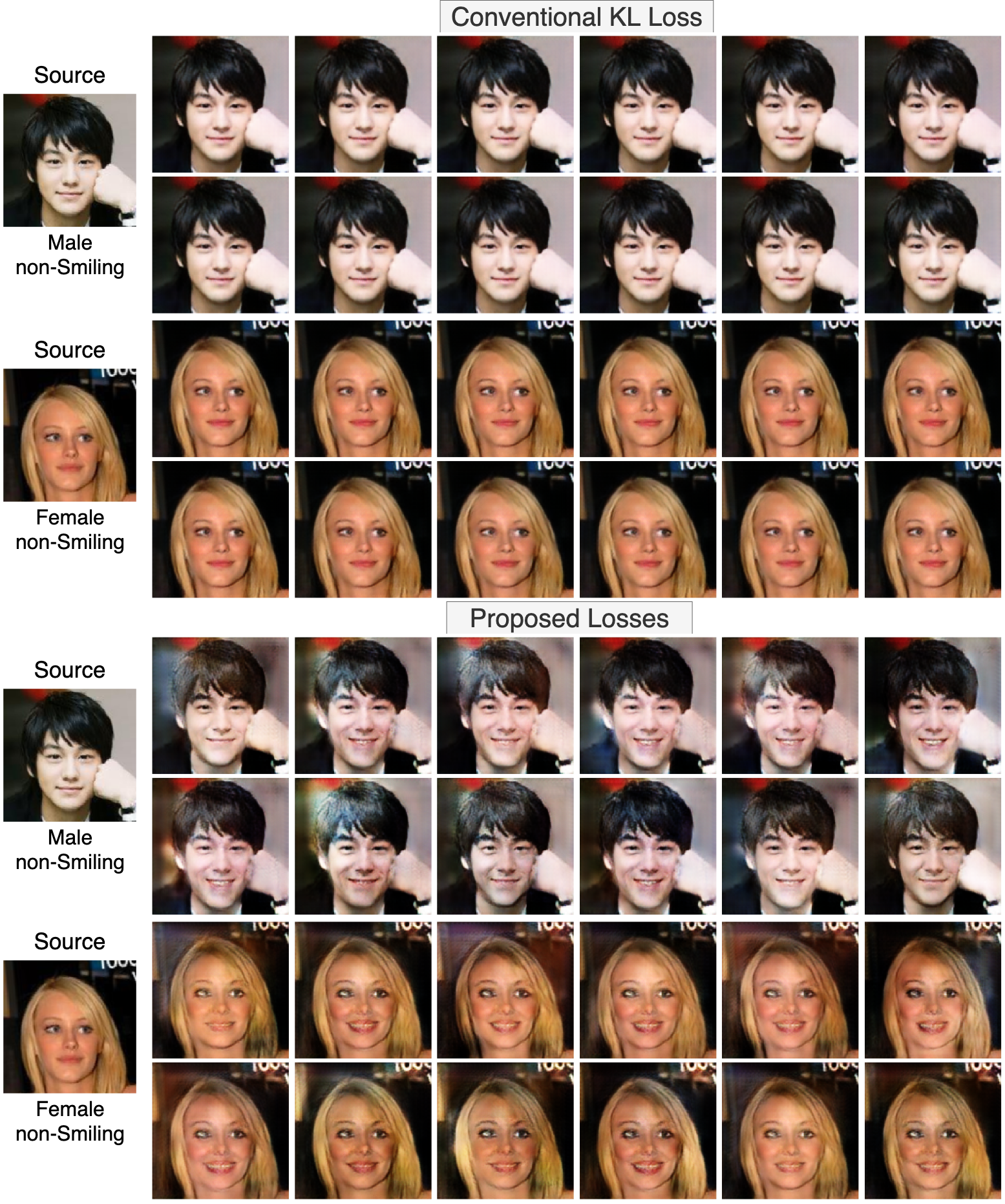}}
\caption{\textit{Comparison between Conventional KL Loss and Proposed Losses (Images)}: 
Every 2 rows show the several outputs of the generator with the same source image and with different style vectors.
While the upper 2 blocks represent the result of the employment of conventional KL loss as a restriction,
the other 2 blocks are the outputs from the proposed 3 losses.
Apparently, the latter outputs were better in terms of diversity.
}
\label{fig:result_diversity_image}
\end{center}
\vskip -0.2in
\end{figure}

Fig.\ref{fig:result_diversity_image} shows the smile maker result of both models, where the improvement is self-evident. First, there was almost no change in appearance with conventional KL loss. In contrast, using the proposed losses, diversification was significantly improved, even though it was not perfect due to the background and/or hair color changes. This characteristic can also be found in the evaluation scores of all the feature extractors in the quantitative results, as shown in Table~\ref{table:qr1}. Therefore, it is clear that, in terms of diversity (recall and coverage), the model using the newly proposed losses surpassed that using the conventional KL loss for all conditions.

\begin{table}[h]
\caption{\textit{Quantitative Results (ex.1)}}
\label{table:qr1}
\vskip 0.15in
\begin{center}
\begin{small}
\begin{sc}
\renewcommand{\arraystretch}{1.25}
\begin{tabular}{ll||c|c|c|c}
\toprule
\multirow{2}{6em}{Dataset for Evaluation} & \multirow{2}{6em}{Model} & \multicolumn{4}{c}{Score} \\
& & Precision & Recall & Density & Coverage \\

\midrule 
\multirow{2}{6em}{Random-Initialization} & ConventionalKL & \textbf{0.83086} & 0.00000 & \textbf{0.77012} & 0.00770 \\
                   & ProposedLoss \textit{(ours)}    & 0.83017 & \textbf{0.00002} & 0.68857 & \textbf{0.06161} \\  \hline 
\multirow{2}{8em}{ImageNet\cite{imagenet}} & ConventionalKL & \textbf{0.72528} & 0.00000 & \textbf{0.79562} & 0.01855 \\
                   & ProposedLoss \textit{(ours)}   & 0.66589 & \textbf{1.8750e-06} & 0.65802 & \textbf{0.04345} \\ \hline 
\multirow{2}{6em}{CelebA\cite{celeba}} & ConventionalKL & \textbf{0.98737} & 0.026391 & \textbf{1.00645} & 0.06418 \\
                   & ProposedLoss \textit{(ours)}  & 0.98414 & \textbf{0.198484} & 0.96772 & \textbf{0.21998} \\
\midrule
        
\bottomrule
\end{tabular}
\renewcommand{\arraystretch}{1}
\end{sc}
\end{small}
\end{center}
\vskip -0.1in
\end{table}

\begin{figure}[!h]
\vskip 0.2in
\begin{center}
\centerline{\includegraphics[scale=0.25]{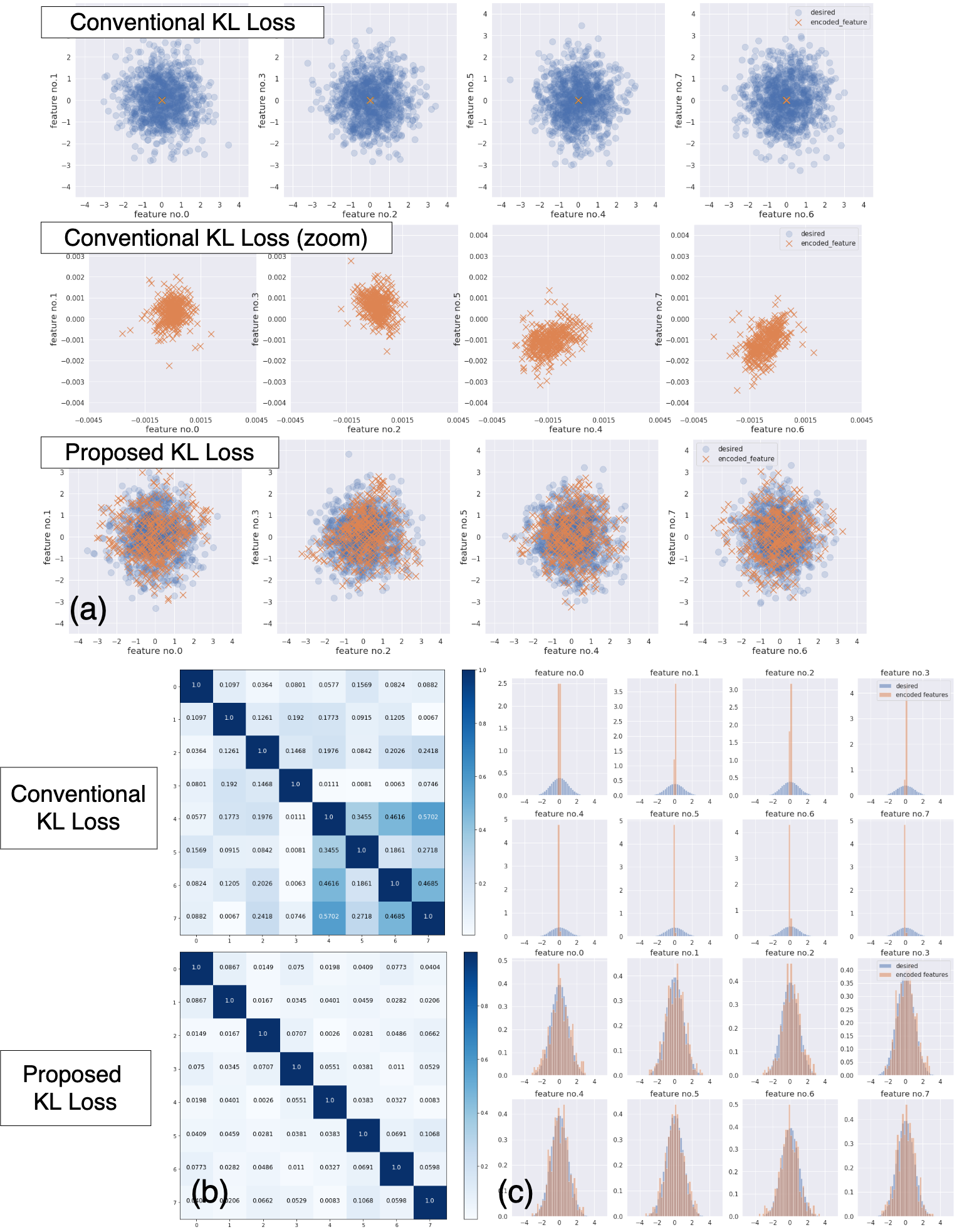}}
\caption{\textit{Comparison between Conventional KL Loss and Proposed Losses in terms of diversity}:
(a) It shows the distribution of encoded features visualized with every 2 features.
The blue circle indicates the desired distribution with the same data size as the encoded features and
the orange cross is the encoded feature.
The 1st row represents the outputs of the encoder with conventional KL Loss.
Since the encoded features were not so visible due to over-restriction, the 2nd-row shows zoomed representations.
The 3rd one is the distribution when the encoder was trained with proposed losses.
Obviously, the last distribution was much better regulated.
(b) It represents the correlation matrix of encoded features in each restriction method.
The upper one is the correlation matrix of the encoder's outputs trained with conventional KL loss,
while the other is that with proposed losses.
It proves that the correlation matrix with the proposed losses is much more similar to that with
the Gaussian distribution.
(c) It shows the histogram of each feature (output) of the encoder.
Each block is from conventional KL loss and the proposed losses from top to bottom.
The histogram from proposed losses is better regulated.
}
\label{fig:result_diversity}
\end{center}
\vskip -0.2in
\end{figure}

This result is primarily due to the distribution of encoded features. Fig.~\ref{fig:result_diversity} shows (a) the distribution of the encoded features, (b) the correlation matrix (absolute value) of the encoded features, and (c) the histogram of the encoded features in each dimension, respectively. As can be seen in Fig.\ref{fig:result_diversity}(a), each of the eight features are paired with the same number of data and displayed along with the desired distribution (Gaussian). Table~\ref{table:mean_std} shows the mean and the standard deviation. Due to the conventional KL loss over-restriction, the encoded features were almost zero compared to the desired outcomes.
In contrast, the encoded features when using the proposed losses were scattered in a manner similar to Gaussian distribution, and the correlation matrix and the histogram were both well-restricted in the proposed losses, as shown in Fig.\ref{fig:result_diversity}(b)(c). The importance of the distribution lies in the regression loss shown in Eq.\ref{eq:regression}, which is designed to encourage a generator to exploit a noise vector as a style vector via the encoder output. Therefore, if the distribution of the encoded features is too far from the desired distribution, it becomes meaningless.

\newpage

\begin{table}[h]
\caption{\textit{Mean and Standard Deviation of the Encoded Features}}
\label{table:mean_std}
\vskip 0.15in
\begin{center}
\begin{small}
\begin{sc}
\renewcommand{\arraystretch}{1.25}
\begin{tabular}{ll||cccccccc}
\toprule
\multirow{2}{6em}{Restriction} & \multirow{2}{*}{} & \multicolumn{8}{c}{Features} \\
& & 1 & 2 & 3 & 4 & 5 & 6 & 7 & 8 \\

\midrule 
\multirow{2}{6em}{Coventional KL Loss} & mean (e-03) & -0.016 &  0.315 &  0.160 &  0.618 & -1.406 & -1.045 & -0.672 & -1.132 \\
                                       & std  (e-03) &  5.090 &  5.178 &  5.158 &  5.964 &  7.592 &  5.781 &  6.542 &  6.491 \\ \hline
\multirow{2}{6em}{Proposed Losses \textit{(ours)}} & mean  & 0.023 &  0.058 &  0.122 &  0.141 & 0.007 & 0.019 & 0.108 & 0.009 \\
                                              & std   & 1.153 &  1.051 &  1.051 &  1.008 & 1.034 & 1.061 & 1.084 & 1.018 \\
\midrule
        
\bottomrule
\end{tabular}
\renewcommand{\arraystretch}{1}
\end{sc}
\end{small}
\end{center}
\vskip -0.1in
\end{table}

\subsection{Style Restriction}
In the second experiment, we examined diversity restrictions by comparing four different models, which were defined with the conditions presented in Table~\ref{table:ex2}. As mentioned above, two concepts were employed to enhance the final result in terms of diversity and fidelity, including regression loss in identity image and UnrolledGAN~\cite{unrolledGAN}. The value “multiple” in the “discriminator” condition means we employed multiple discriminators used in SingleGAN, which matched the number of discriminators with the number of classes (four for this experiment), while “single” indicates that we used the same single discriminator used in StarGAN, which contains an auxiliary classifier. 

\begin{table}[!htb]
\caption{\textit{Hyper-Parameters and conditions (ex.2)}}
\label{table:ex2}
\vskip 0.15in
\begin{center}
\begin{small}
\begin{sc}
\renewcommand{\arraystretch}{1.1}
\begin{tabular}{c||cccc}
\toprule
Symbol or Condition & Model 1 & Model 2 & Model 3 & Model 4 (ours) \\
\midrule
$\lambda_{idt\_reg} $ & 0.5 & 0.5 & 0.5 & 0.5 \\
$k $ & 5 & 5 & 5 & 5\\
restriction (Gaussian) & proposed & proposed & proposed & proposed \\
Discriminator & multiple & single & single & single \\
Encoder'input for class & True & True & False & False \\
\multirow{2}{*}{\centering pretraining} & \multirow{2}{*}{\centering no-pretraining} & \multirow{2}{*}{\centering no-pretraining} & \multirow{2}{*}{\centering no-pretraining} & \multirow{2}{6em}{\centering classification (CelebA)}\\
& & & & \\
\bottomrule
\end{tabular}
\renewcommand{\arraystretch}{1}
\end{sc}
\end{small}
\end{center}
\vskip -0.1in
\end{table}

\begin{figure}[!htb]
\vskip 0.2in
\begin{center}
\centerline{\includegraphics[scale=0.28]{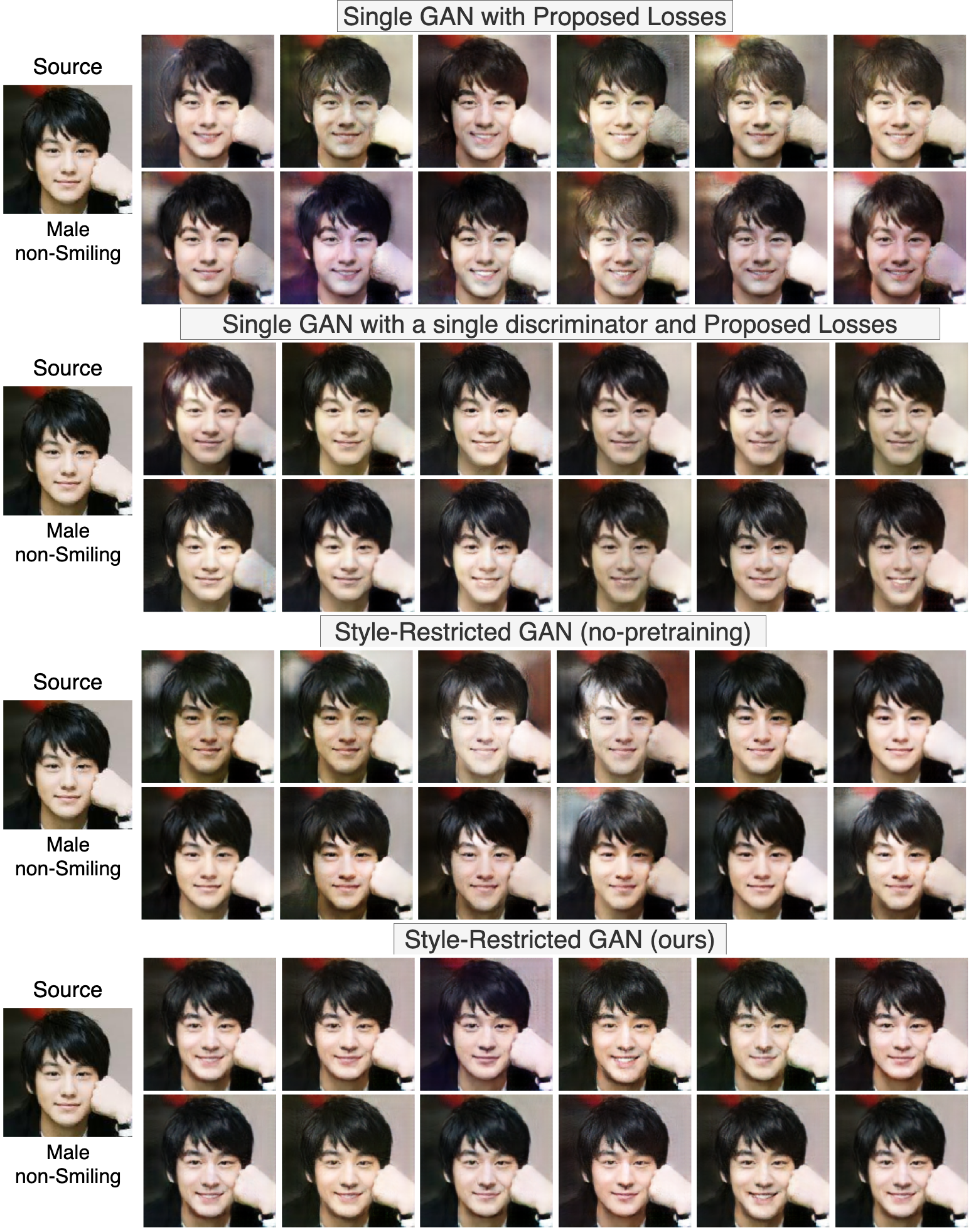}}
\caption{\textit{Comparison among the models designed for style-restriction (male)}:
It shows the results of the (smile) generator for male with different training conditions 
defined by the discriminator and the encoder (Table.\ref{table:ex2})
Each 2 rows is the generator's outputs constructed from the same source image and different style vectors.
As a result, although all models were capable of different translation with different style vectors,
only our model (the last one) was able to focus on the class-related characteristics.
}
\label{fig:result_restriction_male}
\end{center}
\vskip -0.2in
\end{figure}

\begin{figure}[!htb]
\vskip 0.2in
\begin{center}
\centerline{\includegraphics[scale=0.28]{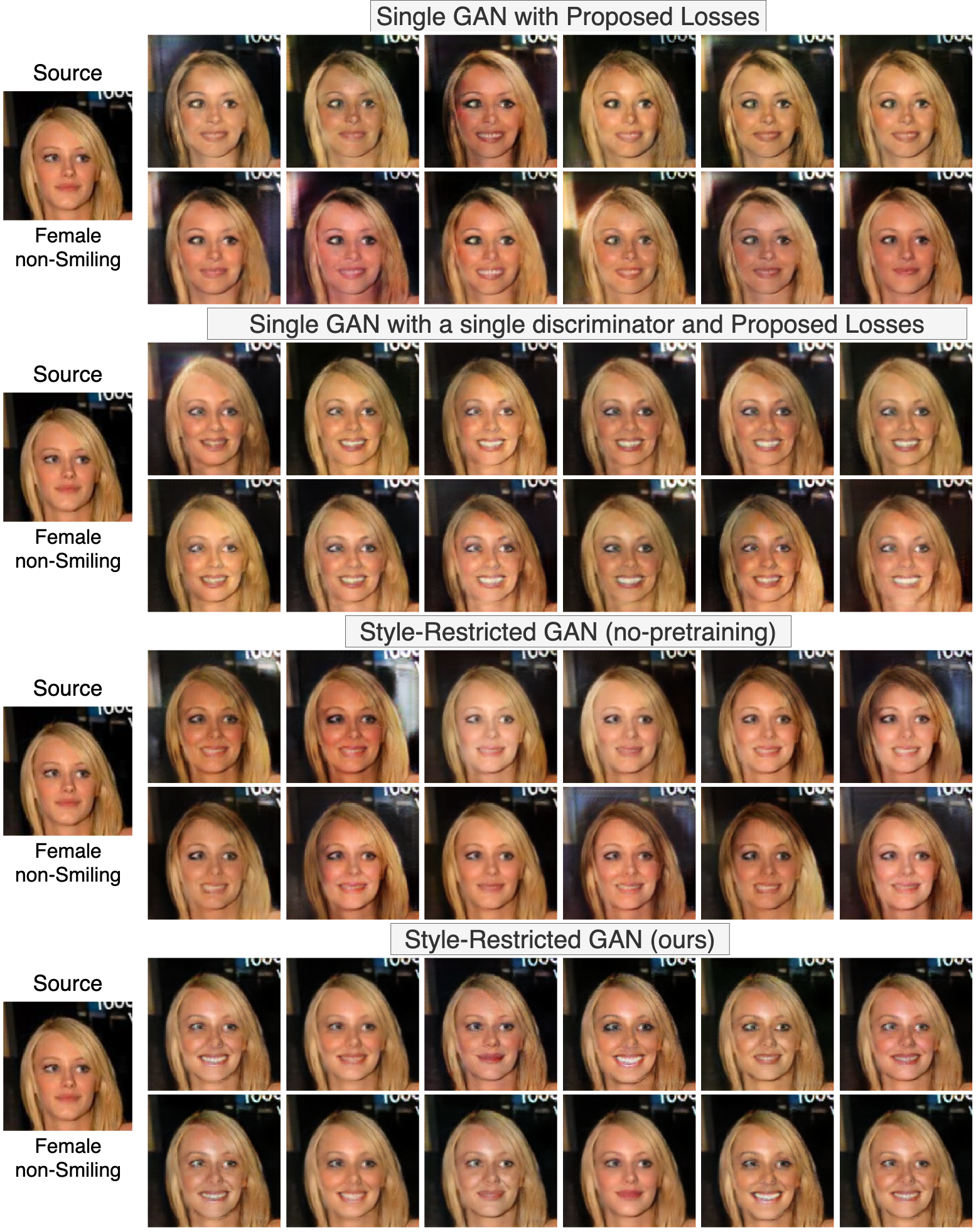}}
\caption{\textit{Comparison among the models designed for style-restriction (female)}:
It shows the results of the (smile) generator for female with different training conditions 
defined by the discriminator and the encoder (Table.\ref{table:ex2})
Every 2 rows is the generator's outputs constructed from the same source image and different style vectors.
As a result, although all models were capable of different translation with different style vectors,
only our model (the last one) was able to focus on the class-related characteristics.
}
\label{fig:result_restriction_female}
\end{center}
\vskip -0.2in
\end{figure}

Figs.\ref{fig:result_restriction_male}~and~\ref{fig:result_restriction_female} show the results of the non-smiling-to-smiling translation for males and females, respectively. Although the two models extended by SingleGAN were both successful in diversifying class-related features, they also diversified features that were unrelated to the class, such as the lighting and the background, hair, and skin colors. For the non-pretrained Style-Restricted GAN (SRGAN), only the class-unrelated characteristics were diversified, so it is reasonable to presume that the class input for the encoder has a relationship with the diversity of class-related features. When SRGAN was pretrained with the classification task, it successfully diversified only the features that were relevant to the classification, which were changes to the smile intensity and eye make-up thickness.

\begin{figure}[!htb]
\vskip 0.2in
\begin{center}
\centerline{\includegraphics[scale=0.28]{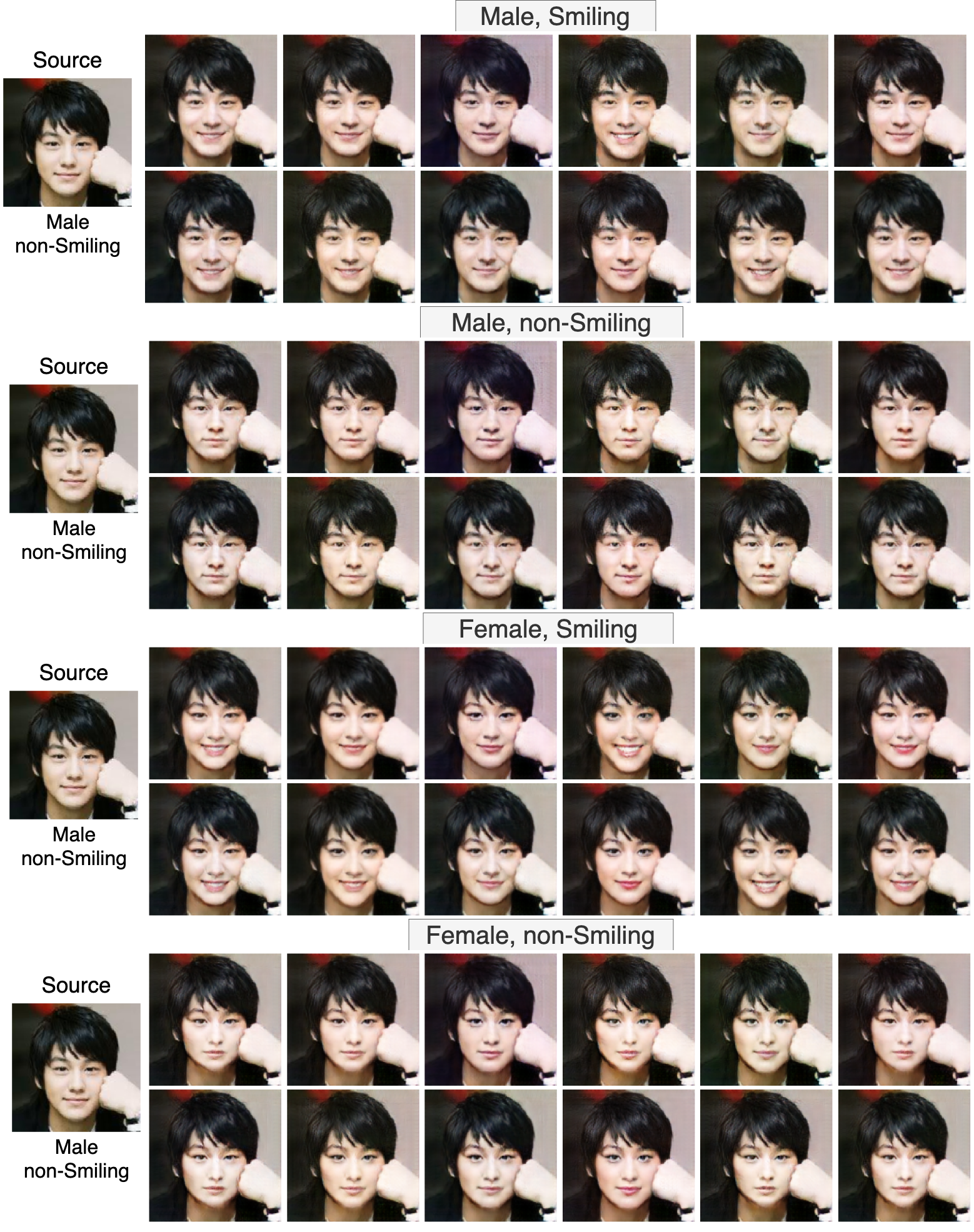}}
\caption{\textit{Additional Results of our proposed model (male 1)}:
It shows additional results of Style-Restricted GAN.
The first 2 rows is outputs of "male, smiling" with different styles and the rest of them are either 
"male, non-smiling", "female, smiling", or "female, non-smiling".
\newline
\newline
}
\label{fig:result_ours_male1}
\end{center}
\vskip -0.2in
\end{figure}

\begin{figure}[!htb]
\vskip 0.2in
\begin{center}
\centerline{\includegraphics[scale=0.28]{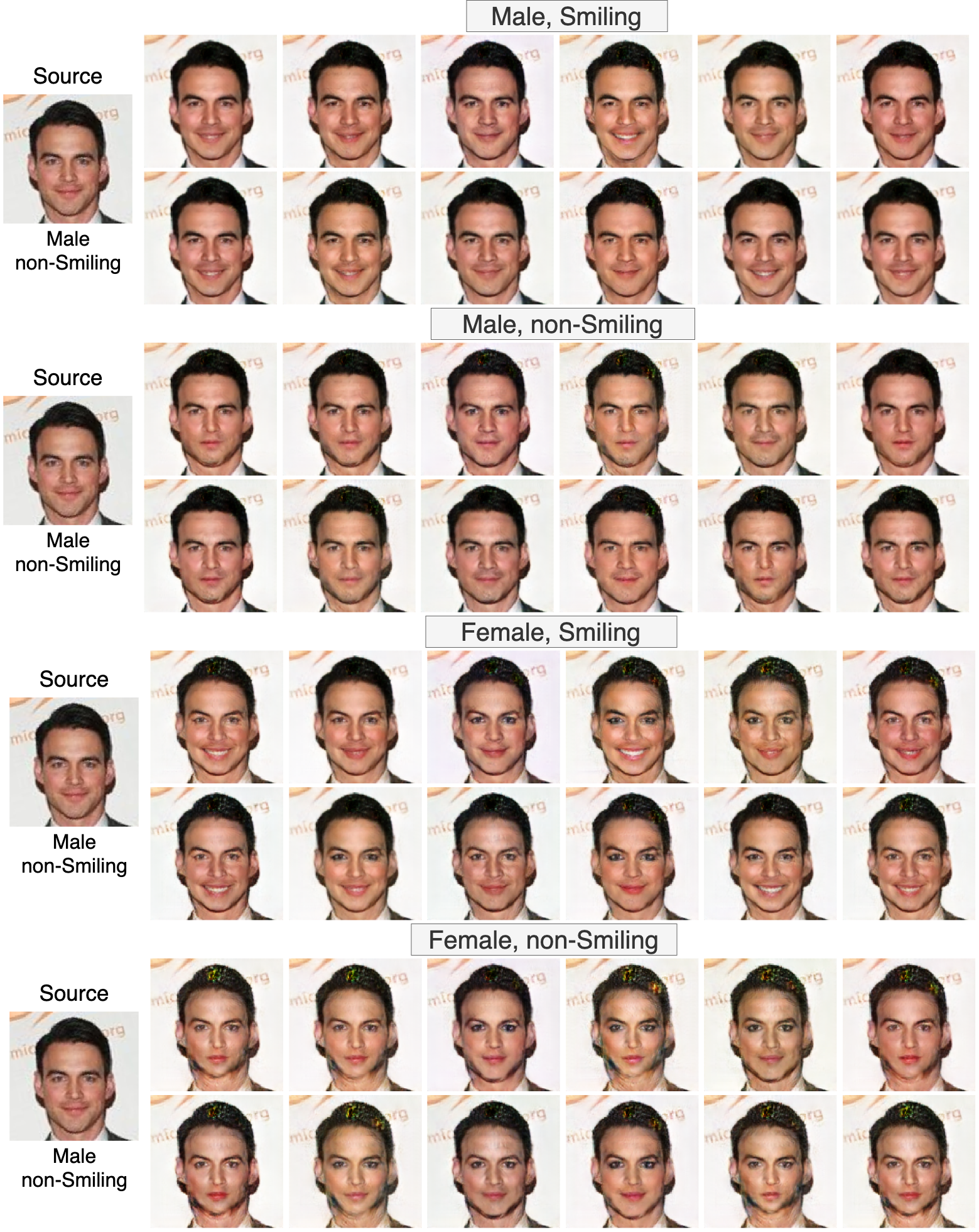}}
\caption{\textit{Additional Results of our proposed model (male 2)}:
It shows additional results of Style-Restricted GAN.
The first 2 rows are outputs of "male, smiling" with different styles and the rest of them are either 
"male, non-smiling", "female, smiling", or "female, non-smiling".
\newline
\newline
}
\label{fig:result_ours_male2}
\end{center}
\vskip -0.2in
\end{figure}

\begin{figure}[!htb]
\vskip 0.2in
\begin{center}
\centerline{\includegraphics[scale=0.28]{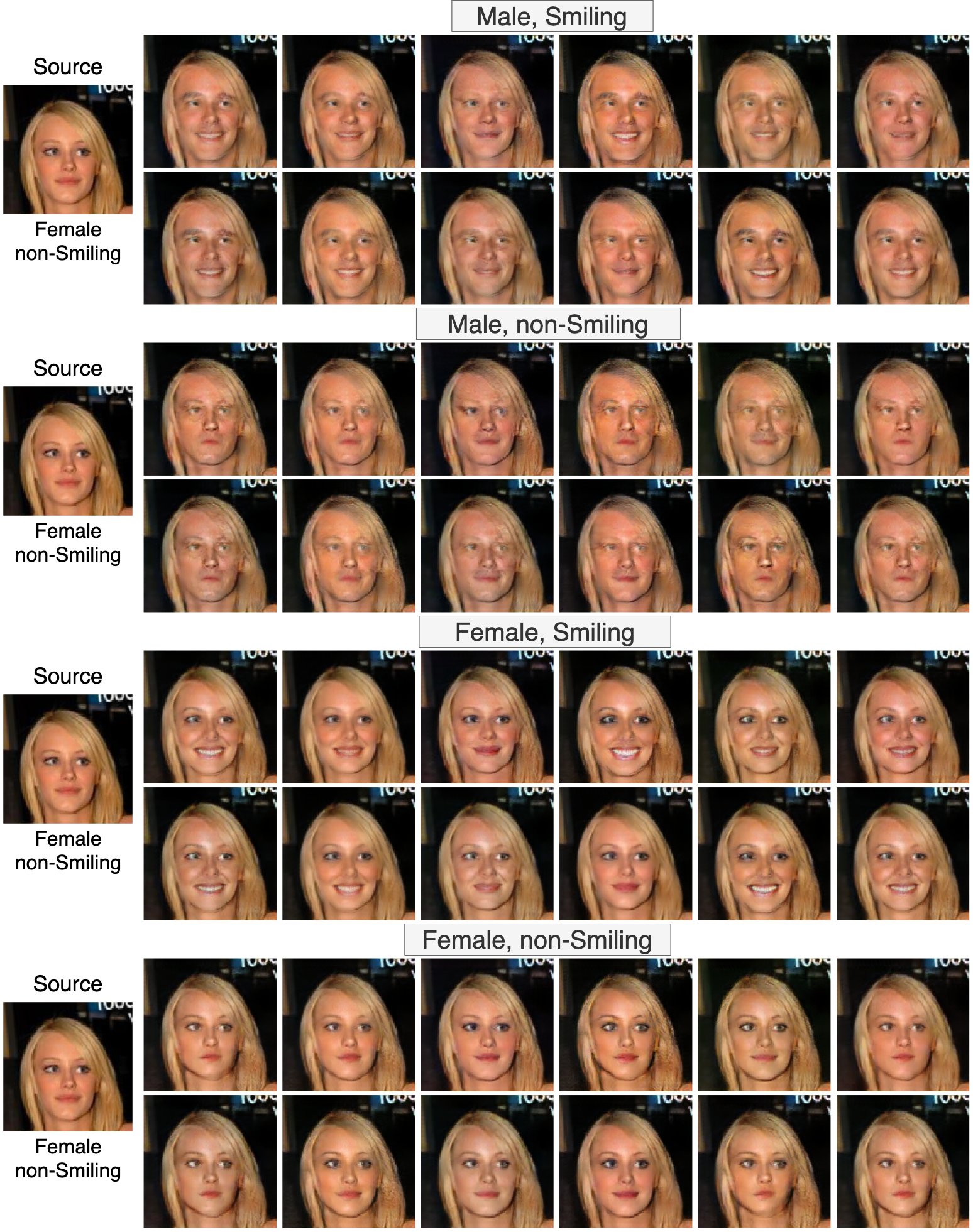}}
\caption{\textit{Additional Results of our proposed model (female 1)}:
It shows additional results of Style-Restricted GAN.
The first 2 rows are outputs of "male, smiling" with different styles and the rest of them are either 
"male, non-smiling", "female, smiling", or "female, non-smiling".
\newline
\newline
}
\label{fig:result_ours_female1}
\end{center}
\vskip -0.2in
\end{figure}

\begin{figure}[!htb]
\vskip 0.2in
\begin{center}
\centerline{\includegraphics[scale=0.28]{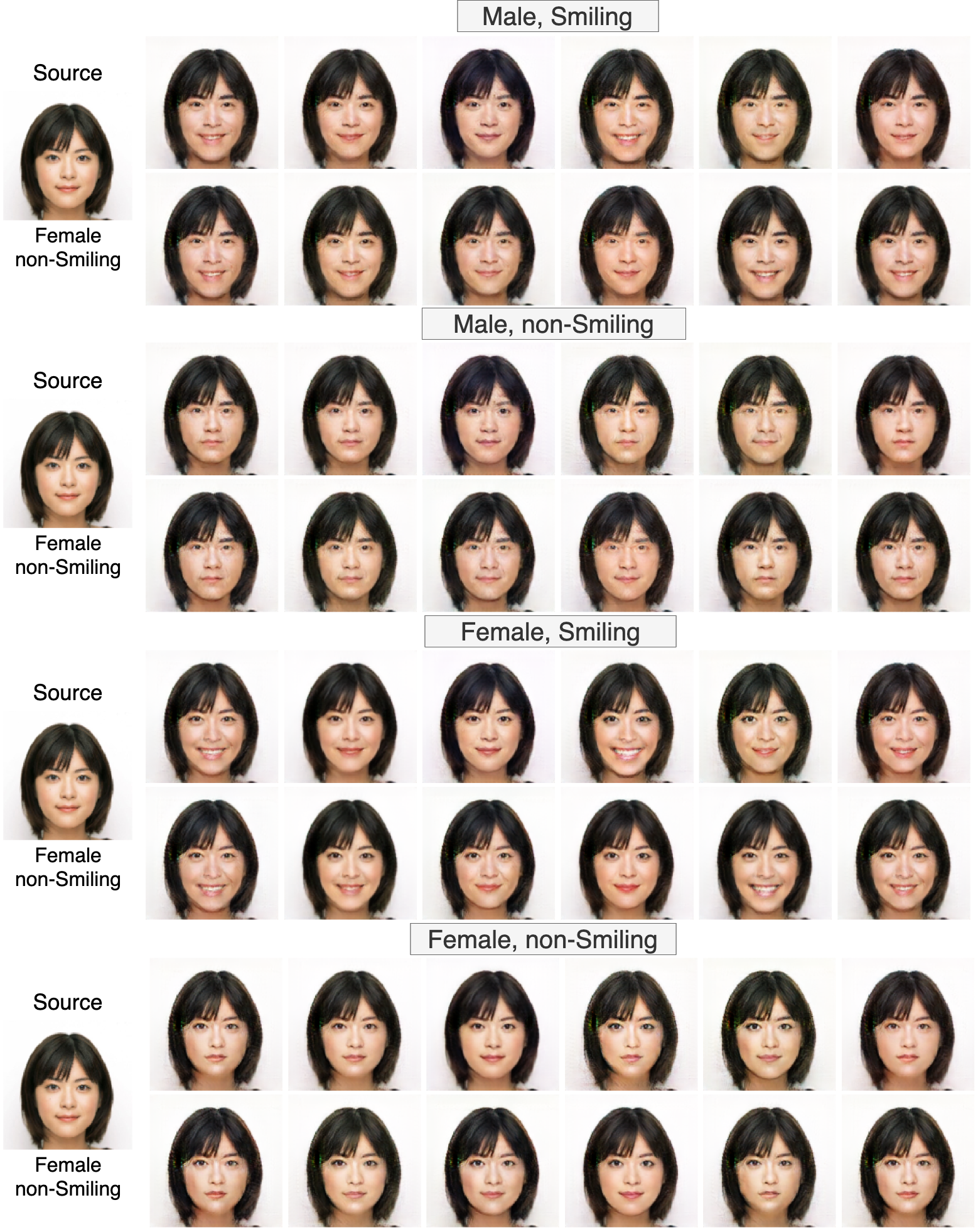}}
\caption{\textit{Additional Results of our proposed model (female 2)}:
It shows additional results of Style-Restricted GAN.
The first 2 rows are outputs of "male, smiling" with different styles and the rest of them are either 
"male, non-smiling", "female, smiling", or "female, non-smiling".
\newline
\newline
}
\label{fig:result_ours_female2}
\end{center}
\vskip -0.2in
\end{figure}

When it comes to the quantitative result with precision, recall, density, and coverage (see Table~\ref{table:qr2}), the diversity score (recall and coverage) obtained with SRGAN (no-pretraining) in random-initialization, which is intuitively able to evaluate genuine diversity, showed the best results. However, this was not the case with the feature extractor pretrained with the face dataset (CelebA), which allowed us to observe class-related diversity. When focusing on the score of the SRGAN pretrained with the classification task, we found that it produced the highest score in both recall and coverage with the CelebA dataset. Additionally, while SingleGAN’s score is almost equal to the SRGAN score in the feature extractor with CelebA dataset, the score with the random-initialized model is nearly 10 times higher than that in SRGAN (pretraining). This indicates that, in comparison with the other translation model, SRGAN (pretraining) improved the \textit{exclusively} class-related diversity. As for the fidelity score (precision and density), for most evaluation models, the score in SRGAN (pretraining) is slightly lower than that in other models. This is because fidelity and diversity scores are likely to be in a trade-off relationship~\cite{dc}. Other qualitative results of Style-Restricted GAN are shown in Figs.\ref{fig:result_ours_male1}~to~\ref{fig:result_ours_female2} at the end of this paper.

\begin{table}[htb]
\caption{\textit{Quantitative Results (ex.2)}}
\label{table:qr2}
\vskip 0.15in
\begin{center}
\begin{small}
\begin{sc}
\renewcommand{\arraystretch}{1.25}
\begin{tabular}{ll||c|c|c|c}
\toprule
\multirow{2}{6em}{Dataset for Evaluation} & \multirow{2}{6em}{Model} & \multicolumn{4}{c}{Score} \\
& & Precision & Recall & Density & Coverage \\

\midrule 
\multirow{4}{6em}{Random-Initialization} & SingleGAN                 & \textbf{0.88711} & 0.00013 & \textbf{0.95987} & 0.11573 \\
                                         & SingleGAN (single D)      & 0.86702 & 0.00011 & 0.89132 & 0.07532 \\
                                         & SRGAN (no-pretraining)    & 0.85179 & \textbf{0.00167} & 0.81889 & \textbf{0.19212} \\
                                         & SRGAN (pretraining, \textit{ours}) & 0.87241 & 0.00000 & 0.92179 & 0.01288 \\ \hline 
\multirow{4}{8em}{ImageNet \cite{imagenet}}              & SingleGAN                 & 0.64997 & \textbf{3.1250e-06} & 0.61623 & \textbf{0.05162} \\
                                         & SingleGAN (single D)      & 0.71701 & 6.2500e-07 & 0.78738 & 0.03944 \\
                                         & SRGAN (no-pretraining)    & 0.67583 & 0.00000 & 0.65013 & 0.05034 \\
                                         & SRGAN (pretraining, \textit{ours}) & \textbf{0.74455} & 0.00000 & \textbf{0.87552} & 0.03441 \\\hline 
\multirow{4}{6em}{CelebA \cite{celeba}}                & SingleGAN                 & 0.98556 & 0.25243 & \textbf{0.98043} & 0.25747 \\ 
                                         & SingleGAN (single D)      & 0.98564 & 0.11649 & 0.96463 & 0.15268 \\
                                         & SRGAN (no-pretraining)    & \textbf{0.98653} & 0.15895 & 0.97914 & 0.17612 \\
                                         & SRGAN (pretraining, \textit{ours}) & 0.98492 & \textbf{0.25309} & 0.97811 & \textbf{0.29953} \\

\midrule
        
\bottomrule
\end{tabular}
\renewcommand{\arraystretch}{1}
\end{sc}
\end{small}
\end{center}
\vskip -0.1in
\end{table}

\section{Conclusion}
In this paper, we proposed Style-Restricted GAN and proved its capabilities in terms of diversification enhancement and restriction 
using both qualitative and quantitative evaluation method.
In other words, it also proved the importance of the encoded features being well-regulated 
in terms of both distribution and class-relation.
To accomplish this, we first experimented with three newly proposed losses (batch KL loss, correlation loss, and histogram imitation loss) against the conventional KL divergence loss in order to restrict the distribution of the encoded features. The results obtained with the proposed losses were much better in mirroring the results of Gaussian distribution. For the diversification restriction or style restriction, the results obtained with the employment of an encoder pretrained in the classification task significantly improved changes to class-related features without changing features that were unrelated to the target class. In our future work, we intend to make use of this model to apply to other domains such as speech data or just simply other facial expressions.

\newpage


\bibliographystyle{unsrt}
\bibliography{references}

\end{document}